\documentclass[10pt,twocolumn,letterpaper]{article}

\usepackage{cvpr}              

\usepackage{graphicx}
\usepackage{amsmath}
\usepackage{amssymb}
\usepackage{booktabs}
\usepackage{multirow}
\usepackage{makecell}
\usepackage{float}
\usepackage{comment}
\usepackage{xspace}
\usepackage{wrapfig}
\usepackage{bm}
\usepackage{caption}
\usepackage{subcaption}

\usepackage[pagebackref,breaklinks,colorlinks]{hyperref}

\usepackage[capitalize]{cleveref}
\crefname{section}{Sec.}{Secs.}
\Crefname{section}{Section}{Sections}
\Crefname{table}{Table}{Tables}
\crefname{table}{Tab.}{Tabs.}

\def\cvprPaperID{9712} 
\def\confName{CVPR}
\def\confYear{2022}

\begin{document}

\title{DoubleField: Bridging the Neural Surface and Radiance Fields for High-fidelity Human Reconstruction and Rendering}

\author{
    Ruizhi Shao$^1$, Hongwen Zhang$^1$, He Zhang$^2$, Mingjia Chen$^1$, Yanpei Cao$^3$, Yu Tao$^1$, Yebin Liu$^1$ \\
    $^1$Tsinghua University \ \ \ $^2$Beihang University \ \ \ $^3$Kuaishou Technology
}

\maketitle


\begin{abstract}
We introduce DoubleField, a novel framework combining the merits of both surface field and radiance field for high-fidelity human reconstruction and rendering. Within DoubleField, the surface field and radiance field are associated together by a shared feature embedding and a surface-guided sampling strategy. Moreover, a view-to-view transformer is introduced to fuse multi-view features and learn view-dependent features directly from high-resolution inputs. With the modeling power of DoubleField and the view-to-view transformer, our method significantly improves the reconstruction quality of both geometry and appearance, while supporting direct inference, scene-specific high-resolution finetuning, and fast rendering. The efficacy of DoubleField is validated by the quantitative evaluations on several datasets and the qualitative results in a real-world sparse multi-view system, showing its superior capability for high-quality human model reconstruction and photo-realistic free-viewpoint human rendering. Data and source code will be made public for the research purpose.
\end{abstract}

\section{Introduction}
The surface fields~\cite{Park_2019_CVPR_deepsdf,mescheder2019occupancy,chen2019learningimplicit} and the radiance fields~\cite{mildenhall2020nerf, zhang2020nerf++} have recently emerged as promising solutions for geometry modeling~\cite{saito2019pifu, saito2020pifuhd, zheng2020pamir, hong2021stereopifu} and texture rendering~\cite{yu2020pixelnerf, peng2020neural,zheng2021structured} of 3D human in an implicit and continuous manner, respectively. 
However, their limitations become apparent when considering simultaneous geometry and appearance reconstruction, not to say under sparse multi-view settings. 
Specifically, the surface fields~\cite{saito2019pifu,hong2021stereopifu,zheng2021deepmulticap,zins2021learning} separate the geometry learning from appearance learning and thus block the joint finetuning ability for more detailed geometry and rendering results. 
Moreover, the radiance fields~\cite{mildenhall2020nerf,peng2020neural,suo2021neuralhumanfvv,peng2021animatable,kwon2021neural} entangle the learning of geometry and appearance in an implicit manner without effective mutual constraints, leading to inconsistent geometry reconstruction and relatively low training efficiency. 
Despite the representations, the feature fusion strategy also dominates the final reconstruction quality when deploying the algorithms under multi-view setups, especially in the real-world systems.
Even with high-resolution images as input, the limited representation power of features (feature map or feature volume)~\cite{saito2020pifuhd,peng2020neural} as well as the calibration and the geometry inference errors (especially for real captured data) will significantly deteriorate the detail reconstruction performance due to multi-view inconsistency for current implicit field based methods~\cite{saito2019pifu,zheng2021deepmulticap,peng2020neural}.

\begin{figure}
    \centering
    \includegraphics[width=1.0\linewidth]{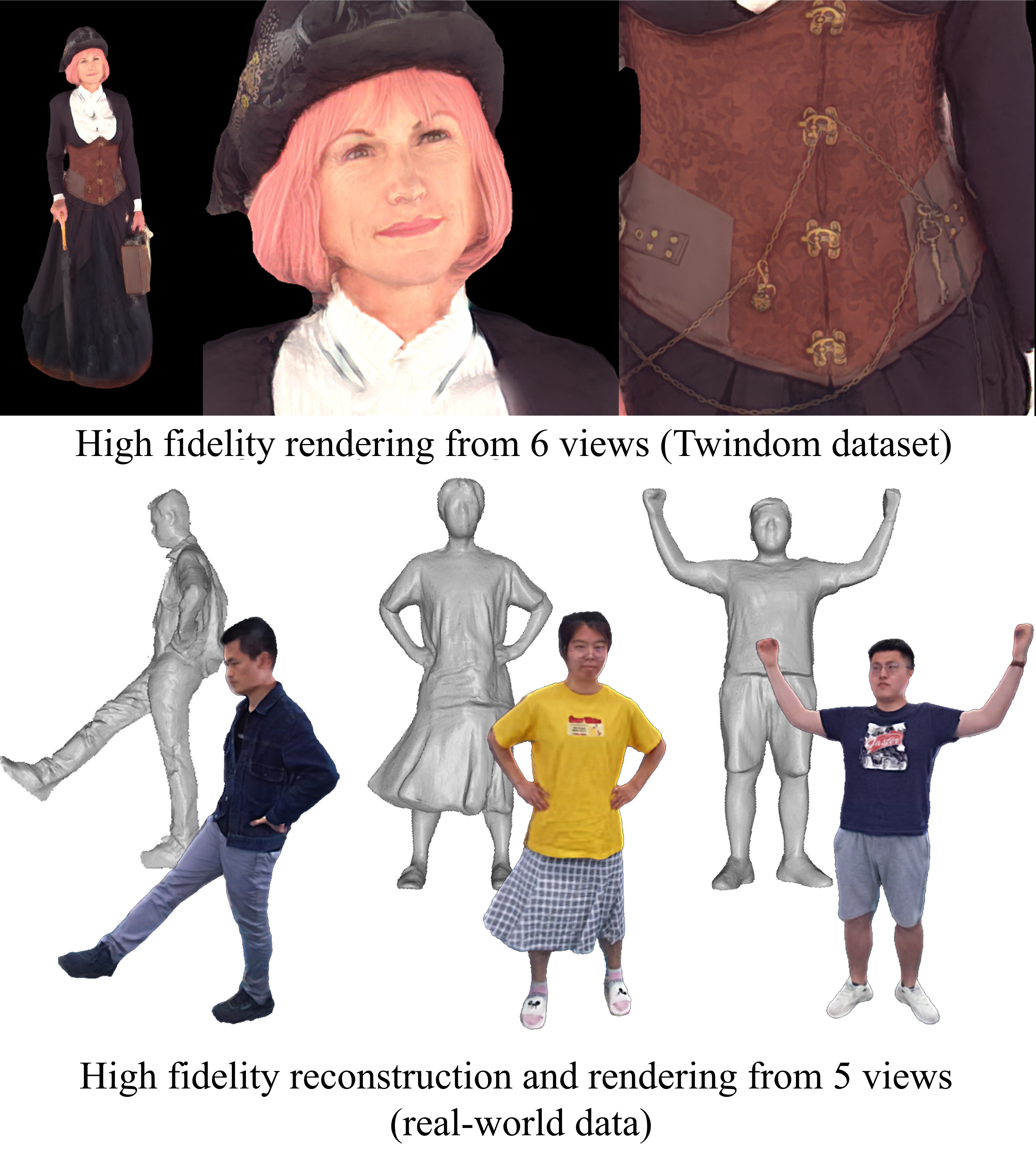}
    \vspace{-0.5cm}
    \caption{
    Given sparse multi-view RGB images, our method achieves high-fidelity human reconstruction and rendering.
    }
    \vspace{-0.5cm}
\end{figure}

To overcome the limitations above for achieving high-quality 3D human reconstruction from sparse-view setups, we propose a novel DoubleField framework (to effectively bridge the surface and radiance fields and enable a shared learning space for both geometry and radiance reconstruction) and a view-to-view transformer (to build self attention between multi-view inputs and cross attention between the input views and the query viewpoints for multi-view feature fusion). 
Specifically, for DoubleField, we build associations between the surface and radiance fields by using a feature embedding shared by these fields in the network architecture and a surface-guided sampling strategy.
Such a shared learning space allows the surface and radiance fields be benefited from each other.
On the one hand, the surface field imposes a geometry constraint to the radiance field and encourages a more consistent density distribution for neural rendering.
On the other hand, the radiance field enables more geometry details in the surface field via differentiable rendering.
Moreover, the surface-guided sampling strategy disentangles the geometry component from the appearance modeling, so that DoubleField has a faster learning process while improving the reconstruction and rendering performances.

When deploying DoubleField with multi-view inputs, we propose a view-to-view transformer to build a self attention between multi-view inputs, and more importantly a cross attention between the input views and the query viewpoints.
We achieve this by adopting an encoder-decoder architecture in our view-to-view transformer.
Specifically, the encoder aims to fuse multi-view features while the decoder aims to produce view-dependent features based on the learned attention between the query view and all input views.
Thanks to the attention learning ability of the transformer, the multi-view inconsistency issue is alleviated in our method, as the attention in the transformer handles the relationships between the input and the query views and is more robust to the geometry inference and calibration errors in real-world multi-view setups.
Besides, the view-to-view transformer also enables our method to utilize the original high-resolution images.
By taking the raw RGB values into accounts, the view-to-view transformer can directly learn the view-dependent features from high-resolution images and contribute to high-fidelity rendering performances.

In comparison with existing approaches~\cite{saito2019pifu,zheng2021deepmulticap,peng2020neural} built upon surface and radiance fields, DoubleField not only improves the reconstruction quality of both geometry and appearance but also has the capability to eliminate the prerequisite SMPL fitting in previous methods~\cite{peng2020neural} and even handle loose clothing (e.g., long dress) .
More importantly, benefiting from the ability to leverage large dataset, DoubleField can fully utilize the priors in the large scale human scan dataset and achieve direct inference and fast finetuning for high-resolution free viewpoint rendering. 
In summary, Our contributions in this work are: 1) a DoubleField framework (a shared double embedding and a surface-guided sampling strategy) to combine the merits of both surface and radiance fields for sparse multiview human reconstruction and rendering; 
2) a view-to-view transformer to fully utilize ultra-high-resolution image inputs in an efficient manner; 3) our method achieve state-of-the-art performance on both geometry reconstruction and texture rendering of human performances using sparse-view inputs. 


\section{Related Work}

\paragraph{Neural implicit field} Recently, neural implicit fields have emerged as powerful representations for geometry reconstruction and graphics rendering.
Compared with the traditional explicit representations, such as meshes, volumes, and point clouds, neural implicit fields encode 3D models via neural networks that directly map 3D locations or viewpoints to the corresponding properties of occupancy~\cite{mescheder2019occupancy, chen2019learningimplicit}, SDF~\cite{Park_2019_CVPR_deepsdf}, volumes~\cite{lombardi2019neuralvolumes}, and radiance~\cite{mildenhall2020nerf} \etc. Conditioned on spatial coordinates rather than discrete voxels or vertices, neural implicit field is continuous, resolution-independent, and more flexible, which enables higher quality surface recovery and photo-realistic rendering. For geometry reconstruction, methods based on surface fields~\cite{saito2019pifu, saito2020pifuhd, xu2019disn} can generate detailed models from one or few images, and the high-fidelity geometry is achieved using local implicit field~\cite{jiang2020local, chabra2020deeplocalshape}. For graphics rendering, methods based on implicit field are suitable for differentiable rendering \cite{liu2020dist, yariv2020multiviewidr, jiang2020sdfdiff, sitzmann2019srn,mildenhall2020nerf}. Among them, the recently proposed NeRF~\cite{mildenhall2020nerf} has made significant progress in novel view synthesis and photo-realistic rendering, which inspires many derivative methods~\cite{yu2020pixelnerf, martin2020nerfwild, Schwarz20neurips_graf, wang2021ibrnet, liu2020nsvf, pumarola2020dnerf} and applications.
Recently, there are also concurrent works~\cite{oechsle2021unisurf,wang2021neus,yariv2021volume} combine surface field and radiance field in an explicit manner and demonstrate promising results for case-specific learning and inference. 
However, extending them to large scale human scan dataset training for general inference is not straightforward. Explicitly building a clear numerical relationship between two fields is also limited to represent solid, non-transparent surfaces. 
In contrast, our DoubleField framework combines these two fields at the feature level in an implicit manner so that we can naturally incorporate pixel-aligned features and learn geometry prior from the large scale dataset during training. 
Our implicit combination is also more suitable to handle general and complicated human cases such as hair, semitransparent skirts, and thin clothes \etc.

\begin{figure*}[t]
    \centering
    \begin{subfigure}[]{0.33\textwidth}
      \centering
      \includegraphics[width=0.9\columnwidth]{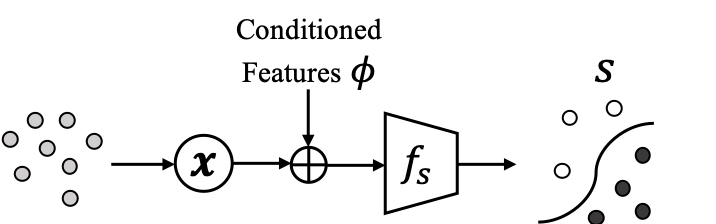}
      \caption{}
      \label{fig:repres_occ}
    \end{subfigure}
    \begin{subfigure}[]{0.33\textwidth}
      \centering
      \includegraphics[width=0.9\columnwidth]{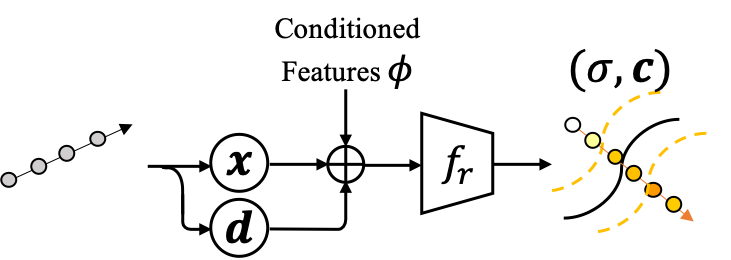}  
      \caption{}
      \label{fig:repres_nerf}
    \end{subfigure}
    \hfill
    \begin{subfigure}[]{0.33\textwidth}
      \centering
      \includegraphics[width=0.9\columnwidth]{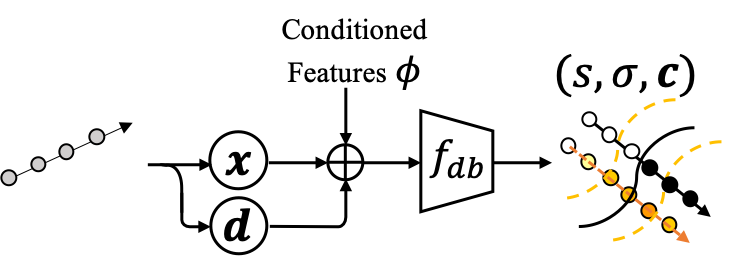}
      \caption{}
      \label{fig:repres_db}
    \end{subfigure}
    \hfill
    \caption{Comparison of different neural field representations. \textbf{(a)} Neural surface field in PIFu~\cite{saito2019pifu}. \textbf{(b)} Neural radiance field in PixelNeRF~\cite{yu2020pixelnerf}. \textbf{(c)} The proposed DoubleField. The joint implicit fucntion $f_{db}$ bridges the surface field and the radiance field.}
    \vspace{-3mm}
    \label{fig:representation}
\end{figure*}

\paragraph{Human reconstruction}
Lately, there are numerous efforts devoted to capturing template-based human body from monocular or multi-view cameras at different levels, including shape and pose~\cite{huang2017towards,liang2019shape,kanazawa2018end,zhang2021pymaf,tian2022hmrsurvey}, and cloth surface~\cite{de2008performance,vlasic2008articulated,gall2009motion,dou2013scanning,xu2018monoperfcap}.
Limited by the representation ability, these methods typically have low-quality results for both geometry and appearance recovery. Moreover, it is also difficult for those template-based algorithms to handle topology changes. 
Other approaches to high-quality human reconstruction require extremely expensive requirements such as dense viewpoints~\cite{joo2018total,wu2020multineuralhuman, liu2009continuous, liu2009point} or even controlled lighting~\cite{collet2015high, guo2019relightables}. Recently, implicit fields~\cite{huang2018deep, zheng2021deepmulticap, saito2019pifu} enable detailed geometry reconstruction from sparse views. 
Based on sparse RGB-D cameras, the high-fidelity geometry reconstruction can be also achieved in real-time~\cite{yu2021function4d, guo2017real, xu2019unstructuredfusion, su2020RobustFusion}. 
Very recently, Peng \etal~\cite{peng2020neural} propose to learn a neural radiance field with the guidance of a predefined template (\ie, SMPL~\cite{loper2015smpl}) and achieve promising results on novel view synthesis from dynamic sequences. However, their method assumes the availability of an accurate estimation of the body template. Moreover, the simultaneous reconstruction of high-fidelity geometry and appearance from sparse-view input remains very challenging for existing solutions. Our work exploits a new path for high-quality geometry reconstruction and high-fidelity human rendering without the need of body templates.

\paragraph{Transformer}
Vaswani et al.~\cite{vaswani2017attention} proposed Transformer, the first sequence transduction model based entirely on attention.
The efficacy of Transformer is recently shown in a wide range of NLP and CV problems~\cite{devlin2018bert,dosovitskiy2020image, yuan2021tokens}. 
The attention mechanism, which is the core of transformer, has been proven by numerous literature to capture long-range dependencies~\cite{vaswani2017attention, wang2018non-local}.
Its ability to obtain correlation has applied to many applications such as visual question answering~\cite{kim2018bilinear}, texture transferring~\cite{yang2020learning}, multi-view stereo~\cite{luo2020attention}, hand pose estimation~\cite{huang2020hot}, and human recontruction~\cite{zheng2021deepmulticap}. 
In our work, we apply a view-to-view transformer to capture the correspondences across the multi-view inputs.

\begin{figure*}[htbp]
    \centering
    \includegraphics[width=0.95\linewidth]{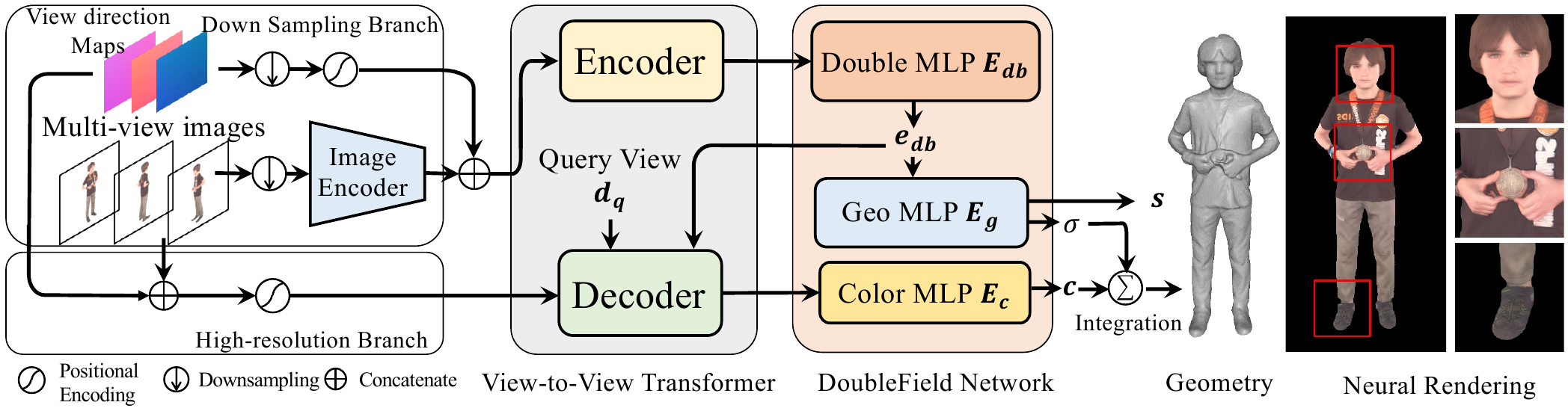}
    \caption{Pipeline of our method. Given sparse multi-view segmented images and the view direction maps, the encoder of our view-to-view transformer fuses low-resolution image features from different viewpoints and output the fused features. The double MLP $E_{db}$ takes the fused features as inputs and produces the double embedding $\bm{e}_{db}$, which will be used to predict the surface field $s$ and the density value $\sigma$ by the geometry MLP. For the prediction of high-fidelity texture, the decoder takes the double embedding $\bm{e}_{db}$, query view direction $\bm{d}$, known views direction $\bm{d}^i$ and the colored encoding $\bm{p}(\bm{x})$ of the ultra-high-resolution images as inputs and produces the texture embedding $\bm{e}_c$ for the prediction of color values $\bm{c}$.}
    \label{fig:pipeline}
\end{figure*}

\section{Preliminary}
Our DoubleField couples the representation power of the neural surface field~\cite{saito2019pifu} and the radiance field~\cite{mildenhall2020nerf,yu2020pixelnerf}. In this section, we give a brief introduction of these two fields.

\paragraph{Neural Surface Field}
The neural surface field represented as the occupancy field~\cite{mescheder2019occupancy,saito2019pifu} is a resolution-independent representation for modeling 3D surface.
As shown in Fig.~\ref{fig:repres_occ}, a surface field can be formulated as an implicit function $f_s$ mapping 3D points $\bm{x}$ to the surface field value $s$, e.g. $f_s(\bm{x})=s: s \in [0, 1]$.
To improve generalization and obtain detailed geometry, PIFu~\cite{saito2019pifu} conditions it on pixel-aligned image features using the following formulation:

\begin{equation}
    f_s(\bm{x}, \phi(\bm{x}, \bm{I})) = s,
\end{equation}
where $\phi(\bm{x}, \bm{I})$ is the image features located at the projection of $\bm{x}$ on the image $\bm{I}$.
PIFu further extends this formulation to reconstruct texture on the surface by predicting RGB color $\bm{c}$ on the points $\bm{x}_c$ satisfied $f_s(\bm{x}_c)=0.5$:  $f_c(\bm{x}_c, \phi(\bm{x}_c, \bm{I})) = \bm{c}$.
Though PIFu provides a straightforward solution for jointly modeling geometry and appearance, it isolates geometry and texture and makes the texture learning space discontinuous, hindering the geometry optimization process under texture supervisions~\cite{niemeyer2020differentiable}.

\paragraph{Neural Radiance Field}
As shown in Fig.~\ref{fig:repres_nerf}, NeRF~\cite{mildenhall2020nerf} represents a scene as a continuous volumetric radiance field $f_r$ of the density $\sigma$ and color $\bm{c}$, which describes geometry and appearance in an entangled form: e.g. $f_r(\bm{x}, \bm{d})=(\sigma, \bm{c})$, where $\bm{d}$ is the viewing direction. Under this formulation, volumetric rendering can be used to synthesize novel view images by integrating along the projection rays:
\begin{equation}
    \hat{C}(\bm{r}(t)) = \int_{t_n}^{t_f}T(t)\sigma(t)\bm{c}(t)dt,
\end{equation}
where $\bm{r}(t)=\mathbf{o}+t\mathbf{d}$ denotes a camera ray with the origin $\mathbf{o}$ and direction $\mathbf{d}$. $T(t)=\exp{(-\int_{t_n}^{t}\sigma(s)ds)}$ tackles with occlusion, and $[t_n, t_f]$ is the pre-defined depth bounds.
To achieve novel view synthesis from only sparse multi-view inputs, PixelNeRF~\cite{yu2020pixelnerf} extends NeRF to leverage pixel-aligned image features in a similar manner to PIFu:
\begin{equation}
    f_r(\bm{x}, \bm{d}, \phi(\bm{x}, \bm{I})) = (\sigma, \bm{c}).
\end{equation}
Since the entangled modeling of density and color brings high flexibility for the training of NeRF, the surface learned in PixelNeRF is inconsistent given only sparse-view inputs, which leads to artifacts such as ghost-like or blurry results in novel view rendering. In addition, the highly flexible nature of the vanilla NeRF makes the training, and finetuning of its derivative solutions~\cite{yu2020pixelnerf,peng2020neural} time-consuming.

\section{Method}

Our method is built on top of the DoubleField network and a view-to-view transformer.
As illustrated in Fig.~\ref{fig:pipeline}, given only sparse-view segmented images with ultra-high resolutions (e.g., 4K), our method can achieve both high-fidelity geometry and appearance reconstruction results without using any human body template.

In this Section, we first introduce our DoubleField network by bridging the surface field and the radiance field in an implicit manner (Sect.~\ref{sec:dbfield}). Based on DoubleField, an efficient view-to-view transformer is designed to leverage high-resolution images and adaptively synthesis photo-realistic rendering results (Sect.~\ref{sec:v2vtrans}).
Our network also supports efficient finetuning to recover high-fidelity geometry and appearance from high-resolution images (Sect.~\ref{sec:imple}).

\subsection{DoubleField Network}
\label{sec:dbfield}
To overcome the limitations of existing neural field representations, we introduce the DoubleField network.
The core of DoubleField consists of a shared embedding and a surface-guided sampling strategy, which connects the surface field and the radiance field so that they can be benefited from each other.

Basically, DoubleField can be formulated as a joint implicit function $f_{db}$ represented by multi-layer perceptrons (MLPs) to fit both the surface field and the radiance field: $f_{db}(\bm{x}, \bm{d})=(s, \sigma, \bm{c})$.
Besides, DoubleField is also conditioned on pixel-aligned images features $\phi(\bm{x}, \bm{I})$. Specifically, as shown in Fig.~\ref{fig:repres_db}, given the query point $\bm{x}$, viewing direction $\bm{d}$ and images features $\phi(\bm{x}, \bm{I})$, our DoubleField network $f_{db}$ learns a shared double embedding and predicts the surface field $s$, the density field $\sigma$ and the texture field $c$ simultaneously. Our DoubleField network is composed of a shared MLP (the \textit{Double MLP} $E_{db}$) for double embedding $e_{db}$ and two individual MLPs 
(the \textit{geometry MLP} $E_g$ and the \textit{texture MLP} $E_c$) 
for the surface field and the radiance field prediction, as illustrated in Fig.~\ref{fig:pipeline}.
Overall, our DoubleField network can be formulated as:
\begin{equation}
\begin{split}
\mathbf{e}_{db} = E_{db}(\gamma(\bm{x}), &\phi(\bm{x}, \bm{I})),\\
    (s, \sigma) = E_g(\mathbf{e}_{db}),~&~c = E_c(\mathbf{e}_{db},\bm{d}),\\
f_{db}(\bm{x}, \bm{d}, \phi(\bm{x}, \bm{I})) & = (s, \sigma, \bm{c}),
\end{split}
\label{eq:doublefield}
\end{equation}
where $\gamma(\bm{x})$ is the positional encoding of $\bm{x}$,
$E_g$ is a \textit{geometry MLP} for the prediction of occupancy $s$ in the surface field and the density $\sigma$ in the radiance field, while $E_c$ is a \textit{texture MLP} for prediction of the color $c$ in the radiance field. Since $s$ and $\sigma$ are two output values of the last layer in the same MLP, such formulation implicitly builds a strong association between the two fields and enables their cooperation at the feature level.

\paragraph{Surface-guided Sampling Strategy}
To further facilitate the relation learned between the two fields and accelerate the rendering process, we make full use of the surface field and propose a surface-guided sampling strategy for DoubleField.
The surface-guided sampling strategy will determine the intersection points in the surface field at first and then perform fine-grained sampling around the intersected surface. Specifically, given camera parameters of the rendering view and the ray $\mathbf{r}=\mathbf{o}+t\mathbf{d}$, a uniform sampling is firstly applied along the ray in the depth bounds $[t_n, t_f]$ with $N_s$ sampling points, and each point is formulated as $\bm{x}_i=\mathbf{o} + t_i\mathbf{d}$. 
We query the surface field value of each point to determine the first intersection position $min\{t_i|\ s(\mathbf{o}+t_i\mathbf{d})\geq0.5\}$ on the surface.
These intersections are then used to guide the sampling at a more fine-grained level by considering the radiance field surrounding the real surface in an interval of $\delta$ with $N_r$ sampling points.

Our surface-guided sampling strategy can emphasize the relation between two fields around the mesh surface which facilitates the training and the finetuning process. Compared with NeRF sampling, our strategy is much fast on account of less sampling points needed for integration.

\subsection{View-to-View Transformer}
\label{sec:v2vtrans}

When applying DoubleField to multi-view inputs, we need to fuse the features from multi-view images.
A straight forward solution is adopting a fusing strategy similar to PIFu~\cite{saito2019pifu} or PixelNeRF~\cite{yu2020pixelnerf}, where the pixel-aligned features will be extracted from the multi-view images and then fused together for DoubleField inference.
Specifically, given image inputs $\{\bm{I}^i\}(i=1,2,...,n)$ from $n$ viewpoints and the corresponding camera parameters, the image features are first extracted by the image encoder. For the query point $\bm{x}$, the pixel-aligned features $\phi^i(\bm{x}, \bm{I}^i)$ on the image $\bm{I}^i$ are first obtained based on the projection of $\bm{x}$.
These pixel-aligned features extracted from the multi-view images are then fused together as $\bm{\Phi}(\bm{x})$:
\begin{equation}
\begin{split}
  \Phi^{i}  & = \oplus(\phi^{i}(\bm{x}, \bm{I}^i), \bm{d}^{i})  \\
  \bm{\Phi}(\bm{x}) & =  \psi(\Phi^{1}, ..., \Phi^{n}),
\end{split}
\label{eq:multiview_fusion}
\end{equation}
where $\oplus(...)$ is a concatenation operator, $\phi^{i}(...)$ is the pixel-aligned features on the $i$-th viewpoint image, $\bm{d}^{i}$ is the viewing direction in the coordinate system of the $i$-th input viewpoint, and $\psi(...)$ is a feature fusion operation such as average pooling~\cite{saito2019pifu} or self-attention~\cite{zheng2021deepmulticap}.
The fused features $\bm{\Phi}(\bm{x})$ can be taken as the conditioned features for DoubleField in Eq.~\ref{eq:doublefield} to predict the corresponding geometry and appearance in the query direction $\bm{d}_q$: $f_{db}(\bm{x}, \bm{d}_q,\bm{\Phi}(\bm{x})) = (s, \sigma, \bm{c})$.

Although the above multi-view feature fusion methods can produce robust and plausible results, they only leverage the relatively low resolution image feature maps.
Moreover, the geometry inference errors and the noises of the calibration in real-world data also significantly limit the quality of the final rendering results.
To overcome this limitation, we propose a view-to-view transformer to directly take the raw RGB values from high-resolution images as input with both self attention and cross attention schemes. 

Specifically, our view-to-view transformer adopts an encoder-decoder architecture that leverages the observations of the point $\bm{x}$ from all input views, and more importantly, the direction $\bm{d}_q$ of the query view to predict the color feature $\bm{e}_c$ for view-dependent rendering.
In this way, our view-to-view transformer not only effectively fuses multi-view features in its \textbf{encoder} but also enables the cross attention between the query view and all the input views in its \textbf{decoder}, which differs from existing transformer-based fusion methods~\cite{zheng2021deepmulticap} that only use the transformer as an encoder for self-attention between input views.
In the following, we present the encoder and decoder of our view-to-view transformer.

\paragraph{Encoder}  
The goal of the encoder is to fuse the geometry features from multi-view inputs.
It adopts the self-attention and feed-forward operation $\psi$ in Eq.~\ref{eq:multiview_fusion} to obtain the fused features $\bm{\Phi}$, which will be fed into the \textit{double MLP} $E_{db}$ for the generation of the double embedding:

\begin{equation}
\begin{split}
  Q^e, K^e, V^e  & = F^e_{Q, K, V}(\phi^1, ..., \phi^n)  \\
  \Phi  & = F^e(Att(Q^e, K^e, V^e))  \\
 \bm{e}_{db} & =E_{db}(\gamma(\bm{x}), \bm{\Phi}),  \\
\end{split}
\label{eq:encoder}
\end{equation}
where $F^e_{Q,K,V}$ denotes the linear layers producing the query, key and value matrices $Q^e, K^e, V^e$, respectively, $F^e$ is the feed-forward layer, and $Att$ is the multi-head attention operation in the transformer.

\vspace{-0.3cm}

\paragraph{Decoder} The goal of the decoder is to produce the view-dependent color embedding $\bm{e}_{c}$ according to the observations from all input views, and the query view direction $\bm{d}_q$.
To leverage the high-resolution information, the decoder takes both low- and high-level observations into account, including the raw rgb $\bm{p}^i$ and double embedding $\bm{e}_{db}$.
Specifically, the process can be formulated as:
\begin{equation}
\begin{split}
  Q^d  & = F^d_Q(\bm{d}_q)  \\
  K^d  & = F^d_K(\bm{d}^1,...,\bm{d}^n)  \\
  V^d  & = F^d_V([\bm{e}_{db},\gamma(\bm{p}^1)],...,[\bm{e}_{db},\gamma(\bm{p}^n)])  \\
  \bm{e}_{c} & = F^d(Att(Q^d, K^d, V^d))
\end{split}
\label{eq:decoder}
\end{equation}
where $F^d_Q,F^d_K,F^d_V$ denote the linear layers producing the query, key and value matrices $Q^d, K^d, V^d$, respectively, $F^d$ is the feed-forward layer. 
Here, similar to the position encoding $\gamma(\bm{x})$, we also map the raw RGB values $\bm{p}^i$ to a higher dimensional space as the colored encoding $\gamma(\bm{p}^i)$ for the learning of high-frequency appearance variations~\cite{2020Fourier}. 

After obtaining the color embedding $\bm{e}_{c}$ from the decoder, the high-resolution color at the point $\bm{x}$ is predicted by the \textit{texture MLP} $E_c$: $\bm{c} = E_c(\bm{e}_{c})$.



\subsection{Training and Finetunning}
\label{sec:imple}

Though our network can leverage high-resolution images as input, the expensive training time cost on such a high-resolution domain is unacceptable.
For a more feasible solution, in implementation we divide the problem into two phases: low-resolution large-scale-dataset pre-training and efficient person-specific high-resolution finetuning.
\vspace{-0.3cm}
\paragraph{Large-Scale Dataset Pre-training}
Our pre-training phase is similar with the training process of PIFu~\cite{saito2019pifu} and PixelNeRF~\cite{yu2020pixelnerf}. 
We collect human models from Twindom\footnote[1]{https://web.twindom.com/} dataset (1,500 for training) and render low-resolution images with the size of $512 \times 512$. 
We adopt the spatial sampling strategy in PIFu~\cite{saito2019pifu} for the learning of geometry, and the proposed surface-guided sampling strategy for the learning of appearance. For the loss of geometry training, we adopt the spatial sampling loss function in PIFu~\cite{saito2019pifu} and the implicit geometric regularization loss (L1 form)~\cite{gropp2020implicit}:
\begin{equation}
\begin{split}
   L_g & = \frac{1}{N_g} \sum_{i=1}^{N_g} \| s(\bm{x_i}) - s^*(\bm{x_i}) \|^2_2 \\
   L_r & = \frac{1}{N_r} \sum_{i=1}^{N_r} \| \nabla s(\bm{x_i}) - n^*(\bm{x_i}) \|_1, 
\end{split}
\label{eq:decoder}
\end{equation}
where $s^*(\bm{x}_i)$ is the ground truth occupancy of $\bm{x_i}$, and $n^*(x_i)$ is the ground truth normal of $\bm{x_i}$. $N_g$ and $N_r$ are the number of sampling points for spatial sampling and geometric regularization, respectively. The regularization loss can further improve the quality of geometry reconstruction without requirement of normal map as input. To obtain the ground truth of normal, we only sample points on the mesh surface when applying regularization loss. And for appearance loss, we adopt the L1 loss between the rendered color and the ground truth color as: 
\begin{equation}
    L_c = \frac{1}{N_c}\sum_{i=1}^{N_c}|\hat{C}(\bm{r}_i) - C^*(\bm{r}_i)|,
\label{color_loss}
\end{equation}
where the rendered color $\hat{C}(\bm{r}_i)$ is obtained using the integration~\cite{mildenhall2020nerf} along the ray $\bm{r}_i$ in the interval around the surface. $C^*(\bm{r}_i)$ is the ground truth color of ray $\bm{r}_i$. $N_c$ is the number of sampling rays. In summary, our final loss can be formulated as:  $L = \lambda_gL_g + \lambda_rL_r + \lambda_cL_c$, where $\lambda$s balance the loss terms.
\vspace{-0.1cm}
\paragraph{Finetuning Phase}
In the finetuning phase, the network takes the ultra-high-resolution images from the sparse multi-view of a specific human as input and finetune the network parameters in a self-supervised manner using differentiable rendering loss. Specifically, We first fix the transformer and the color MLP to finetune geometry for 2000 iterations and then fix the double MLP and the geometry MLP to finetune the color MLP for another 2000 iterations. In each iteration, we randomly select one view as ground truth and regard the other views as input. The only one loss function we used is Eq.~\ref{color_loss} and the learning rate is tune down for stable finetuning performance (1e-6 in finetuning and 1e-5 in pre-training).

\section{Experiment}

\begin{figure}[htbp]
    \centering
    \includegraphics[width=0.95\linewidth]{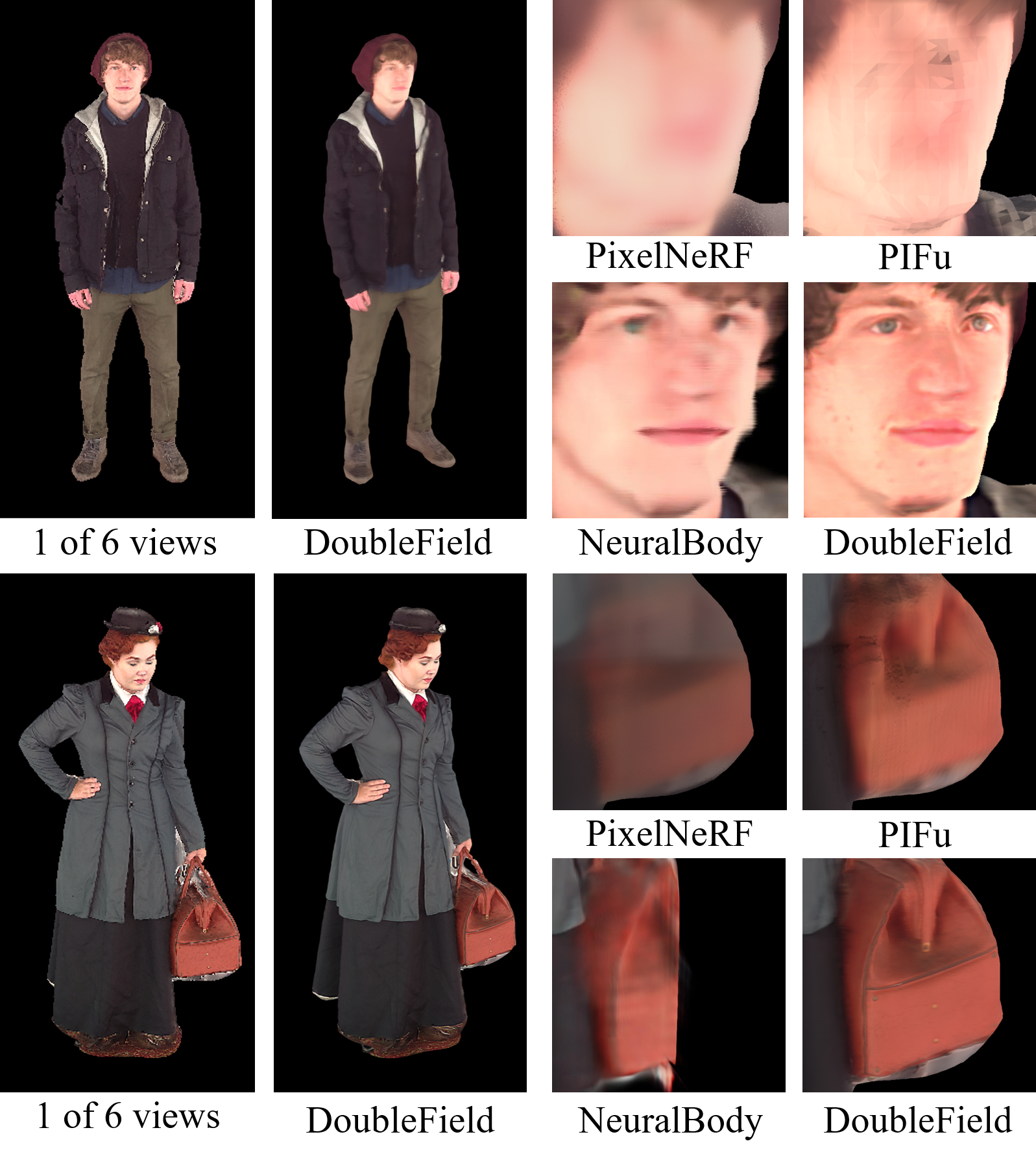}
    \vspace{-0.4cm}
    \caption{Comparison on appearance reconstruction using the Twindom dataset.  PixelNeRF~\cite{yu2020pixelnerf} and our method are finetuned with additional 4,000 iterations. 
    Note that NeuralBody~\cite{peng2020neural} can not handle additional objects which are far away from the human body like handbag. }
    \vspace{-0.35cm} 
    \label{fig:color_cp}
\end{figure}

\begin{figure}[htbp]
    \centering
    \vspace{-0.7cm}
    \includegraphics[width=0.95\linewidth]{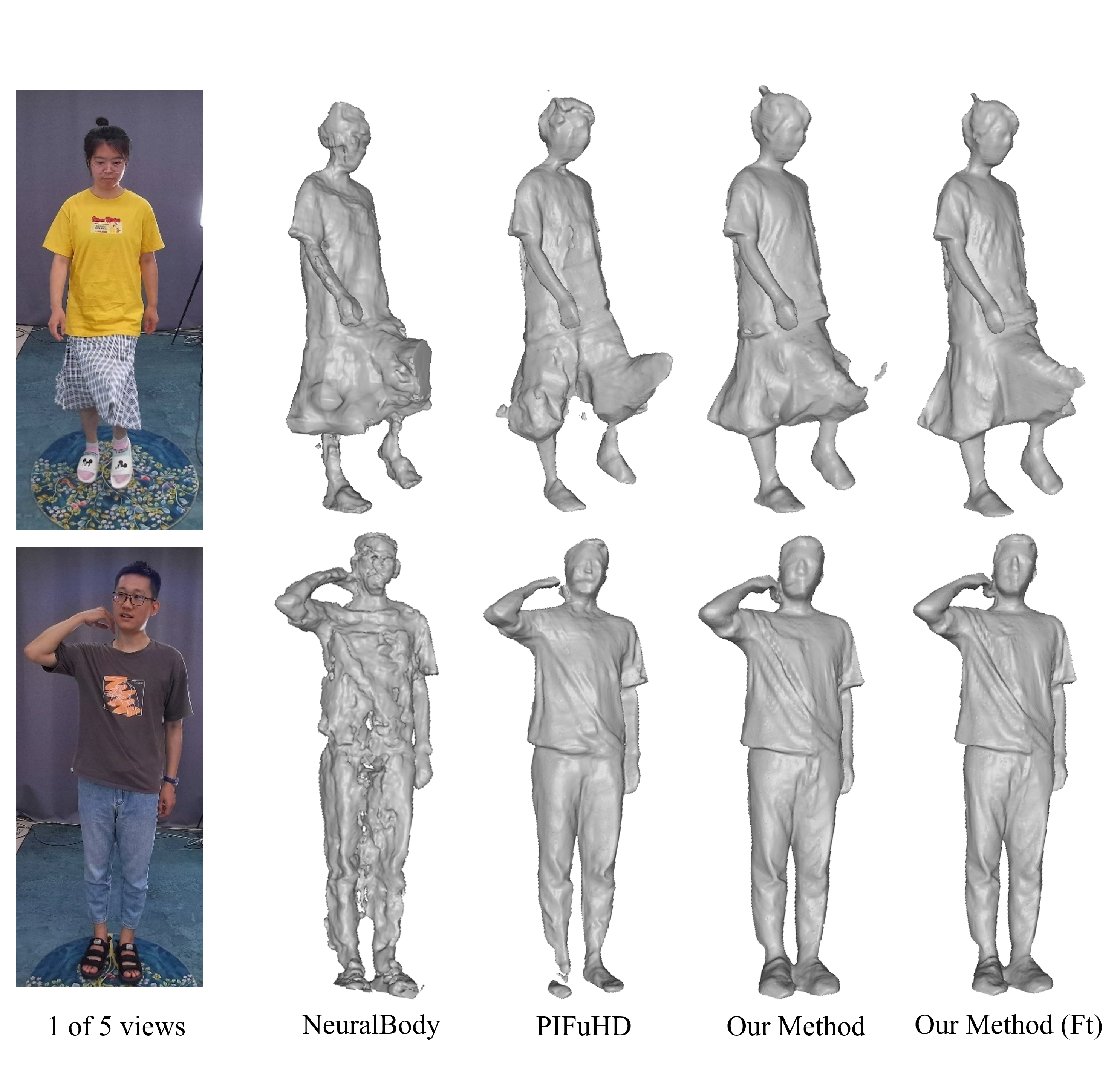}
    \vspace{-0.4cm}
    \caption{Comparisons of geometry reconstruction results using real multi-view image (5 views).}
    \vspace{-0.75cm}
    \label{fig:geometry_cp}
\end{figure}

\begin{table}[htbp]\small
   \centering
   \begin{tabular}{p{2.7cm}|cc|cc}
  \hline
  \multirow{2}*{Method} 
  & \multicolumn{2}{c|}{\makecell[c]{Twindom \\ \small{(6 views Geo.)}}}
  & \multicolumn{2}{c}{\makecell[c]{THuman2.0 \\ \small{(6 views Geo.)}}} \\
         & Chamfer & P2S & Chamfer & P2S \\ \hline
       PIFu~\cite{saito2019pifu}   & 0.754 & 0.716 &  0.710 & 0.613 \\
       PIFu+R   & 0.739 & \textbf{0.699} &  0.697 & 0.606 \\
       PIFuHD~\cite{saito2020pifuhd}   & 0.742 & 0.701 &  0.700 & 0.609 \\
        \makecell[l]{PIFu+DVR~\cite{niemeyer2020differentiable}} & 0.746 & 0.701 & 0.709  & 0.611 \\
       \makecell[l]{PixelNeRF~\cite{yu2020pixelnerf}}   & 0.945 & 0.931 & 0.815  & 0.725 \\
        \makecell[l]{Our Method (w/o Ft)}  &  \textbf{0.737} & 0.700 & \textbf{0.696} & \textbf{0.605} \\  \hline
         \makecell[l]{NeuralBody~\cite{peng2020neural}}& 1.597 & 2.146 &  1.528 & 2.126 \\
       \makecell[l]{PIFu+DVR (Ft)}   & 0.779 & 0.736 &  0.724 & 0.623 \\
       \makecell[l]{PixelNeRF (Ft)} & 1.072 & 1.052 &  0.790 & 0.701 \\
       \makecell[l]{Our Method (Ft)}  & \textbf{0.711} & \textbf{0.690} &  \textbf{0.662} & \textbf{0.589} \\  \hline
  \end{tabular}
  \vspace{-0.3cm}
   \caption{
   Quantitative human geometry reconstruction results. Ft denotes the approaches finetuned with 4,000 iterations.}
  \vspace{-0.3cm}
   \label{tab:dataset_table}
  \end{table}

\begin{table}[htbp]\small
   \centering
   \begin{tabular}{p{2.7cm}|cc|cc}
  \hline
  \multirow{2}*{Method} & \multicolumn{2}{c|}{\makecell[c]{Twindom \\ \small{(6 views Col.)}}}
  & \multicolumn{2}{c}{\makecell[c]{THuman2.0 \\ \small{(6 views Col.)}}} \\
        & PSNR & SSIM & PSNR & SSIM \\ \hline
       PIFu~\cite{saito2019pifu}  & 20.80 & 0.805 & 22.35 & 0.846 \\
        \makecell[l]{PIFu+DVR~\cite{niemeyer2020differentiable}} & 20.65 & 0.804 & 22.17 & 0.843 \\
       \makecell[l]{PixelNeRF~\cite{yu2020pixelnerf}}  & 21.57 & 0.808 & 22.95 & 0.854 \\
        \makecell[l]{Our Method (w/o Ft)} & \textbf{22.95} & \textbf{0.842} & \textbf{24.23} & \textbf{0.880}  \\  \hline
         \makecell[l]{NeuralBody~\cite{peng2020neural}}& 20.69 & 0.808 & 22.65 & 0.862 \\
       \makecell[l]{PIFu+DVR (Ft)}  & 21.62 & 0.812 & 23.08 & 0.855  \\
       \makecell[l]{PixelNeRF (Ft)} & 21.85 & 0.813 & 23.57 & 0.863 \\
       \makecell[l]{Our Method (Ft)} & \textbf{23.56} & \textbf{0.857} & \textbf{25.10} & \textbf{0.905} \\  \hline
  \end{tabular}
  \vspace{-0.3cm}
   \caption{
   Quantitative human rendering results. Ft denotes the approaches finetuned with 4,000 iterations.}
  \vspace{-0.6cm}
   \label{tab:dataset_table}
  \end{table}

\subsection{Experiments on Synthetic Data}

We evaluate our method by using synthetic rendering of multiview images on two high-quality 3D human scan datasets:
1) Twindom dataset (200 for testing),
2) THuman2.0~\cite{yu2021function4d}, a publicly-available high-quality human model dataset (100 for testing).

We compare DoubleField with the state-of-the-art approaches built upon the surface field and the radiance field, including PIFu~\cite{saito2019pifu}, PixelNeRF~\cite{yu2020pixelnerf}, NeuralBody~\cite{peng2020neural}, and PIFuHD~\cite{saito2020pifuhd}.
We also implement DVR~\cite{niemeyer2020differentiable} based on PIFu (denoted as PIFu+DVR) to validate the efficiency of the DoubleField representation and its finetuning ability on unseen data. For fair comparisons, we additionally train PIFu with regularization loss (PIFu+R) and replace the average pooling operation in PIFu~\cite{saito2019pifu}, PIFuHD~\cite{saito2020pifuhd} and PixelNeRF~\cite{yu2020pixelnerf} with self-attention modules for multi-view feature fusion. We retrain these networks with the same training settings and datasets.

\textbf{Comparisons on Geometry Reconstruction.} For the comparison with NeuralBody~\cite{peng2020neural}, we regard NeuralBody as a frame-based method and train it on 6 viewpoint inputs for 15 hours. Due to the expensive training cost, we randomly pick only 50 models from Twindom test dataset and 30 models from THuman2.0 dataset for  NeuralBody evaluation.
We quantitatively evaluate the geometry recovery performance using the point-to-surface distance and the chamfer distance in Table.~\ref{tab:dataset_table}. Our method without finetuning achieves competitive results compared with PIFuHD, PIFu+R and PIFu+DVR. 
After finetuning, our method can further improve the quality of geometry even without ground truth geometry for supervision based on the DoubleField representation. 

\textbf{Comparisons on Appearance Rendering.} To evaluate the appearance rendering performance, we prepare images of 4K resolution rendered from 30 viewpoints, and use images from 6 fixed viewpoints as input and images from other 24 views for evaluation. Quantitative results are shown in Table.~\ref{tab:dataset_table}. Benefiting from the view-to-view transformer and the DoubleField representation, our method achieves high-fidelity rendering. Moreover, our method can support higher quality appearance reconstruction with quick finetuning in 20 minutes (10 minutes for geometry finetuning and 10 minutes for texture and transformer finetuning, 4,000 iterations in total). Moreover, our method generalizes well to scenarios like loose clothes (e.g. long skirts) and object interactions as shown in Fig.~\ref{fig:color_cp}.

\textbf{Ablation Study.} We compare different factors that contribute to our method. As shown in Tab.~\ref{tab:ablation}, compared with the view-to-view transformer and the color encoding, the DoubleField network has the most significant contribution to the final results. Meanwhile, the view-to-view transformer is more effective for achieving multi-view and cross view feature fusion than a simple pooling layer. We also conduct experiments in the high-resolution domain with finetuning. The model of ``Ft w/o HD pixel'' is finetuned  using only low-resolution images (512x512). The performance of such setting is worse than our method but better than the others, demonstrating the ability of our view-to-view transformer to capture correspondences across different views and leverage the high resolution input.

\begin{table}\small
   \centering
   \begin{tabular*}{\linewidth}{p{2.8cm}|cc|cc}
  \hline
  \multirow{2}*{} & \multicolumn{2}{c|}{\makecell[c]{Twindom \\ \small{(6 views Col.)}}}
  & \multicolumn{2}{c}{\makecell[c]{THuman2.0 \\ \small{(6 views Col.)}}} \\
        & PSNR & SSIM & PSNR & SSIM \\ \hline
       average pooling  & 22.53 & 0.826	& 23.89 & 0.870 \\
      w/o DbMLP & 22.01 & 0.818	& 23.42 & 0.866 \\
      w/o CE & 22.89 & 0.831	& 24.11 & 0.874 \\
        Our method (w/o Ft) & 22.95 & 0.842 &	24.23 & 0.880  \\  \hline
        Ft w/o HD pixel &	23.28 & 0.847	& 24.97 & 0.896 \\
       Our method (Ft)  & 23.56 & 0.857 &	25.10 & 0.905 \\   \hline
  \end{tabular*}
  \vspace{-0.3cm}
   \caption{
   Ablation study on Twindom and Thuman2.0 dataset with four settings: Average pooling (use the same multi-view feature fusion in PIFu and PixelNeRF), W/o DbMLP (remove Double MLP and learn two fields separately), W/o CE (removes the color encoding and directly adopt 3-dim RGB), Ft w/o HD pixel (finetune using only low-resolution images). }
  \vspace{-0.5cm}
   \label{tab:ablation}
  \end{table}

\begin{figure*}[htbp]
    \centering
    \includegraphics[width=0.9\linewidth]{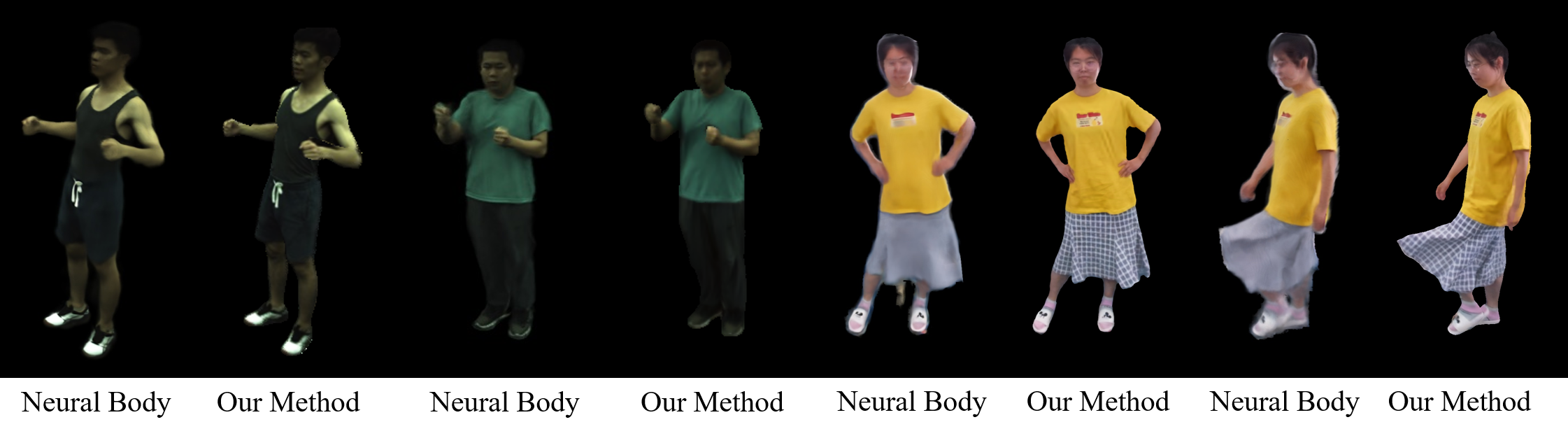}
    \vspace{-0.3cm}
    \caption{Comparisons with NeuralBody. 4 images on the left are from the ZJU-mocap dataset, and 4 images on the right are from a real world multi-view (5 views) system. Each video has 300 frames and we train NeuralBody for 20 hours.}
    \vspace{-0.3cm}
    \label{fig:nb_cp}
\end{figure*}
\begin{figure*}[htbp]
    \centering
    \includegraphics[width=0.85\linewidth]{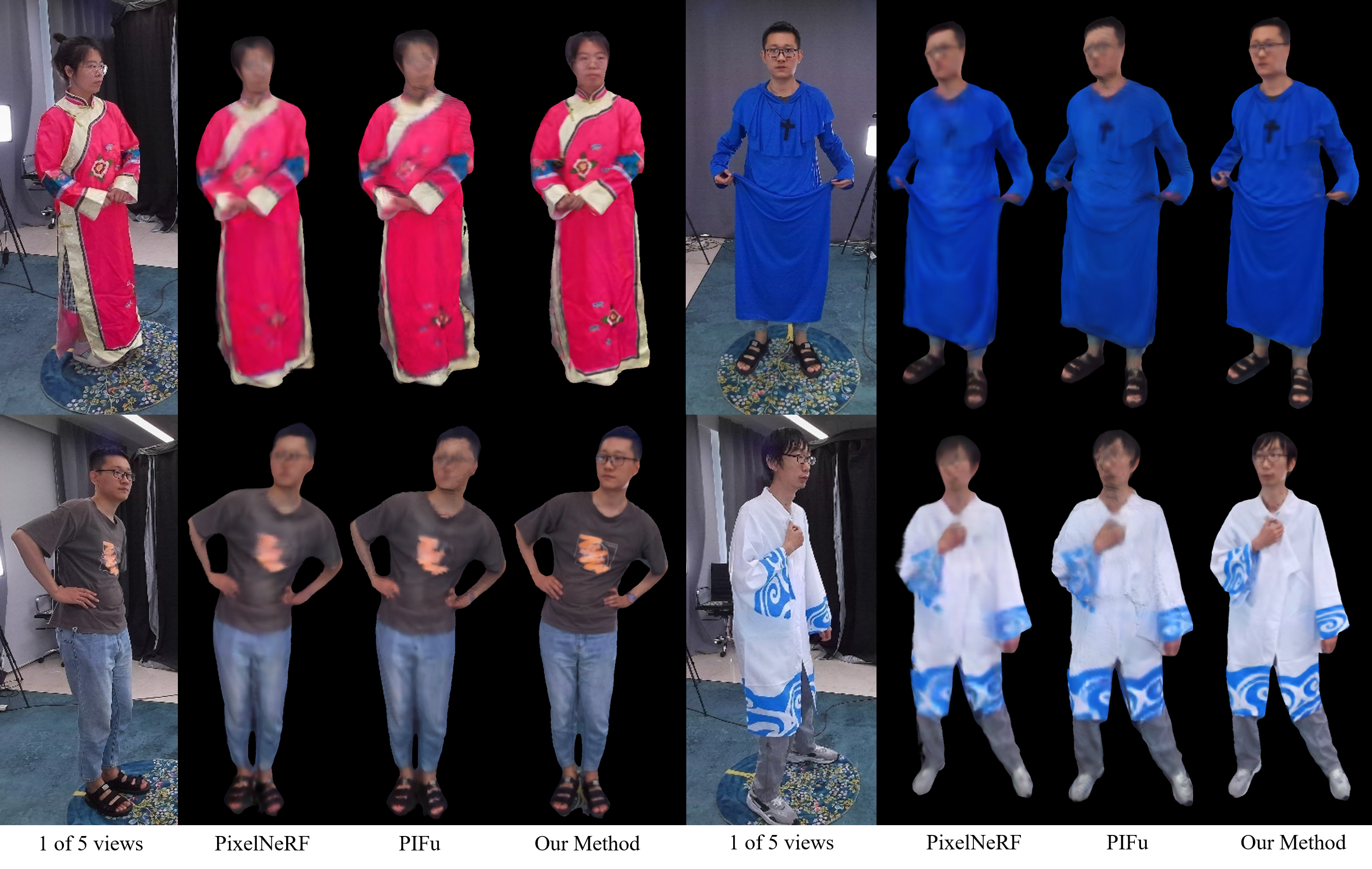}
    \vspace{-0.3cm}
    \caption{Comparisons on real world data under 5-views setting with PixelNeRF\cite{yu2020pixelnerf} and PIFu\cite{saito2019pifu}}
    \label{fig:video_cp}
    \vspace{-0.5cm}
\end{figure*}

\subsection{Results on Real World Multi-view Data}
We evaluate our geometry reconstruction and texture rendering performance using real-world data captured from sparse multi-view cameras (5 views). Fig.~\ref{fig:geometry_cp} compares the qualitative geometry reconstruction results of NeuralBody~\cite{peng2020neural}, PIFuHD~\cite{saito2020pifuhd}, and our method.
Note that our method is finetuned with the multi-view images at one frame, while NeuralBody~\cite{peng2020neural} is trained with the whole mutli-view video sequence as it fails in the geometry reconstruction when only one frame is given.
As show in Fig.~\ref{fig:geometry_cp}, unlike NeuralBody~\cite{peng2020neural}, the surface reconstructed by our method is more consistent and contains more details. The finetuning can further fix some missing parts on the geometry such as holes, which shows that the double MLP has learned to build an implicit association between the two fields. Finally, even without using the normal maps as input, our method produce more accurate results compared with the multi-view extension of PIFuHD.

We further evaluate the rendering quality on the ZJU-mocap dataset~\cite{peng2020neural} and our multi-view system. The results are shown in Fig.~\ref{fig:nb_cp} and Fig.~\ref{fig:video_cp}.
Our method produces more clear rendering results using much less time for network finetuning ($<$ 20 minutes V.S. $>$ 15 hours). Moreover, our method does not rely on human shape prior SMPL~\cite{loper2015smpl} compared with NeuralBody~\cite{peng2020neural} and achieves photo-realistic rendering even under challenging scenarios like swinging skirt, topological changes and loose cloth, which demonstrates the strong generalization capacity of our method to real world data. For more results, please refer to our supplementary video. 



\section{Discussion}
\noindent 
\textbf{Conclusion.} We propose DoubleField to combine the merits of both geometry and appearance fields for human surface reconstruction and rendering under sparse view inputs. In our work, the proposed DoubleField network and view-to-view transformer enable a substantial performance improvement on both geometry reconstruction and texture rendering of human performances. We believe our approach can enlighten the follow-up works in the field of human rendering and reconstruction.

\noindent 
\textbf{Limitations.} 
The proposed pipeline still relies on accurate background image subtraction for Doublefield inference due to the requirement of pixel-aligned image feature extraction. Moreover, our method does not support reconstruction and rendering of multiple character scenarios. 

\noindent 
\textbf{Potential Social Impact.} Our method focuses on free-viewpoint rendering of a human performance and can be used in sport games, movie, virtual reality, tele-presence, etc., which has no obvious negative societal impact.

\noindent 
\textbf{Acknowledgement.}
This paper is sponsored by National Key R\&D Program of China (2021ZD0113503) and the NSFC No. 62125107 and No. 62171255.


{\small
\bibliographystyle{ieee_fullname}
\bibliography{egbib}
}

\end{document}